%% file: main_sangkeum.tex
\documentclass{article}
\usepackage{ijcai26}

\usepackage{times}
\usepackage{soul}
\usepackage{url}
\usepackage[hidelinks]{hyperref}
\hypersetup{
    pdftitle={DEPLOY-RL: Active Boundary Discovery and Conservative Certification for Deployable Reinforcement Learning in Safety-Critical Continuous Processes},
    pdfauthor={Yeojin Jang, Minu Baek, Gihun Gil, Minsung Jung, Beomdo Park, Woohyeon Kwon, Hyeonseok Jang, Junseong Park, Neha Sengar, Andres Saurez, Sangkeum Lee},
    pdfsubject={IJCAI-ECAI 2026 AI4Tech --- Camera-Ready},
    pdfkeywords={Reinforcement Learning, Conformal Prediction, Safety Certification, Active Boundary Discovery, Deployment, Industrial Process Control}
}

\usepackage[utf8]{inputenc}
\usepackage[small]{caption}
\usepackage{subcaption}
\usepackage{graphicx}
\usepackage{booktabs}
\usepackage{amsmath,amssymb,amsthm}
\newtheorem{proposition}{Proposition}
\newtheorem{lemma}{Lemma}
\newtheorem{definition}{Definition}
\usepackage{algorithm}
\usepackage{algpseudocode}
\usepackage{multirow}
\usepackage{xcolor}
\usepackage{siunitx}

\usepackage{enumitem}
\setlist{nosep}

\title{DEPLOY-RL: Active Boundary Discovery and Conservative Certification for Deployable Reinforcement Learning in Safety-Critical Continuous Processes}

\author{
    Yeojin Jang$^1$
    \and
    Minu Baek$^1$
    \and
    Gihun Gil$^1$
    \and
    Minsung Jung$^1$
    \and
    Beomdo Park$^1$
    \and
    Woohyeon Kwon$^1$
    \and
    Hyeonseok Jang$^1$
    \and
    Junseong Park$^1$
    \and
    Neha Sengar$^2$
    \and
    Andres Saurez$^2$
    \And
    Sangkeum Lee$^{1}$\thanks{Corresponding author.}
    \affiliations
    $^1$Hanbat National University, Republic of Korea\\
    $^2$Korea Advanced Institute of Science and Technology, Republic of Korea
    \emails
    \{jyeoj251, bmw5779, minegihun, jmss1101, pbeomdo, mfireon0520, seokchu123, js03093351\}@gmail.com,\\
    \{neha.sengar, asaurez\}@kaist.ac.kr,
    sangkeum@hanbat.ac.kr
}

\begin{document}
\maketitle
\input{results/tables/summary_macros.tex}

\begin{abstract}
	Reinforcement learning (RL) policies often outperform classical controllers in simulation, yet rarely reach production in safety-critical processes.
	The barrier is that there is no principled way to answer ``\emph{Is this policy safe to deploy?}'' with statistical guarantees.
	We introduce \textbf{DEPLOY-RL}, a post-training certification framework built on one key insight: deployment certification requires discovering \emph{where} failures occur (boundary discovery), not measuring \emph{how much} everywhere (uniform reconstruction).
	Our contributions: (1) a contract-coupled acquisition function that concentrates sampling on certification-critical boundaries, achieving $\approx 2\times$ sample efficiency with a semi-empirical ambiguity reduction bound (domain-calibrated convergence guarantee);
	(2) conformal risk control providing finite-sample false-go guarantees ($\le\alpha$) under explicit deployment contracts;
	(3) a three-way decision framework (Deploy/No-Deploy/Abstain) with fail-safe PID/MPC fallback.
	In simulations on papermaking (industrial digital twin) and Tennessee Eastman (public benchmark), DEPLOY-RL achieves \textbf{4.4\% false-go rate} (vs.\ 8.6\% for the best baseline) while retaining \textbf{88.6\% policy coverage}, the only method achieving $<$5\% false-go with $>$85\% coverage among 14 baselines under our evaluation protocol.
\end{abstract}

\section{Introduction}
Reinforcement learning (RL) shows strong simulated performance in industrial process optimization, yet rarely reaches production in safety-critical settings.
The barrier is not algorithmic performance but the absence of \textbf{principled deployment criteria} with \emph{finite-sample risk guarantees}.
Classical PID/MPC is stable but conservative; our framework enables a hybrid strategy: deploy RL when certified safe and fall back to robust PID/MPC otherwise.
This setting differs from offline benchmark selection: the operator must make a single deployment decision for a trained controller under a specified operational stress distribution, where a false positive deployment can cause safety violations while an abstention only incurs fallback cost.

This gap represents a fundamental bottleneck for AI-enabled critical technologies.
Existing safe RL methods (CPO \cite{achiam2017constrained}, Lyapunov-based approaches \cite{chow2018lyapunov}, Recovery RL \cite{thananjeyan2021recovery}) address \emph{training-time} safety but assume deployment.
Black-box validation \cite{corso2021survey,feng2023dense} discovers failures but lacks statistical deployment guarantees.
Recent conformal approaches \cite{lindemann2023safe,luo2024sample} provide prediction intervals but not policy-level certification decisions.
A critical question remains unanswered: \textbf{how should an operator decide whether to deploy a trained policy, defer to fallback control, or reject deployment entirely, all with statistical guarantees?}

\textbf{DEPLOY-RL} is a post-training \emph{certification layer} applicable to any trained policy.
Its key insight is that deployment certification is a \textbf{decision-oriented boundary discovery problem}: instead of uniformly reconstructing failure boundaries, it samples where boundary uncertainty most affects deployment decisions under an explicit contract.
This reframes active testing from ``find the boundary accurately everywhere'' to ``resolve the boundary mass that changes the deploy/no-deploy decision.''

\paragraph{Contributions.}
Our work makes three contributions:
\begin{itemize}
	\item \textbf{Decision-oriented boundary discovery (Section 4).} A \emph{contract-coupled acquisition function} (Eq.~3) prioritizes certification confidence under the deployment contract and gives monotone ambiguity reduction (Proposition~\ref{prop:ambiguity}), yielding $\approx2\times$ sample efficiency empirically.
	\item \textbf{Conformal deployment certification (Section 5).} We cast deployment as \emph{selective conformal risk control} under an explicit contract, with finite-sample false-go bounds (Proposition~\ref{prop:false_go}) and data-separation conditions (Lemma~\ref{lem:no_leakage}).
	\item \textbf{Three-way industrial fallback.} The rule outputs Deploy, No-Deploy, or Abstain; abstention triggers fail-safe PID/MPC fallback.
\end{itemize}
We empirically validate DEPLOY-RL on papermaking (proprietary digital twin) and the Tennessee Eastman benchmark (public) using \emph{deployment-centric} metrics: false-go rate, coverage, and certification cost.

\begin{figure}[!t]
	\centering
	\vspace{-0.4em}
	\includegraphics[width=0.94\columnwidth]{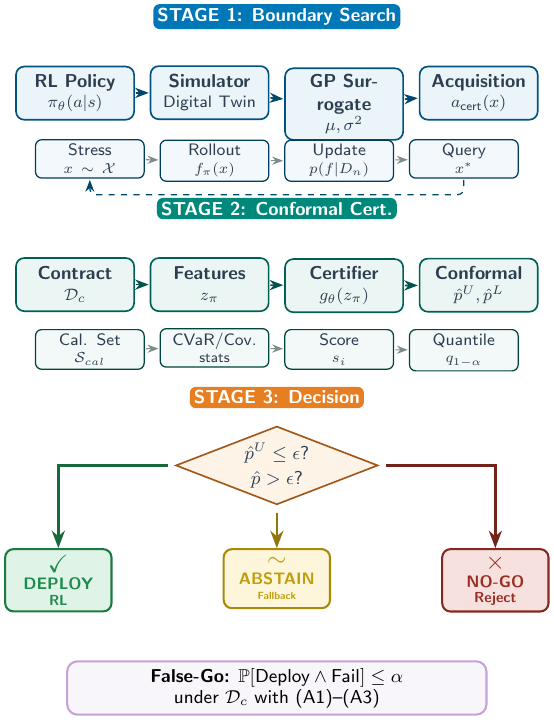}
	\caption{\textbf{DEPLOY-RL pipeline.} Boundary search targets certification-critical regions, conformal calibration produces conservative risk bounds, and the final rule returns Deploy, Abstain, or No-Go with PID/MPC fallback.}
	\vspace{-0.8em}
	\label{fig:pipeline}
\end{figure}

\paragraph{Pipeline overview (Figure~\ref{fig:pipeline}).}
DEPLOY-RL has three stages: contract-aware boundary search, conformal calibration of $\hat{p}^U$, and a Deploy/No-Deploy/Abstain rule with guarantee $\mathbb{P}[\text{Deploy} \land \text{Unsafe}] \le \alpha$ under $\mathcal{D}_{\mathrm{contract}}$.

\section{Related Work}
\paragraph{Safe RL and deployment verification.}
Safe RL methods (CPO \cite{achiam2017constrained}, Lyapunov-based \cite{chow2018lyapunov}, Recovery RL \cite{thananjeyan2021recovery}) address \emph{training-time} safety but assume deployment will occur.
Black-box validation \cite{corso2021survey,feng2023dense}, runtime failure detection \cite{xu2025fail}, probabilistic safety filters \cite{tabbara2025statistically}, and conformal control-barrier approaches \cite{lindemann2023safe,luo2024sample} improve safety evidence, but do not provide post-training policy-level deploy/no-deploy certification with false-go control.

\paragraph{Conformal risk control and selective prediction.}
Classical conformal prediction \cite{vovk2005algorithmic,angelopoulos2021gentle} provides coverage guarantees; recent work extends it to risk control \cite{angelopoulos2024rcps}, decision-oriented conformal methods \cite{fisch2024conformal}, covariate shift \cite{barber2024causal}, feedback adaptation \cite{wang2025conformal}, and adaptive abstention \cite{kumar2025beyond}.
Unlike general conformal risk control, we target \emph{deployment decision correctness}: contract-aligned boundary discovery, data separation despite adaptive sampling, and three-way fallback decisions.
Standard conformal statements certify set coverage, but industrial deployment needs a policy-level statement of the form ``deploy only when false-go probability is controlled.''

\paragraph{Active boundary discovery.}
Level-set estimation \cite{bryan2006active,gotovos2013active}, randomized straddle sampling \cite{inatsu2024active}, stability-informed acquisition \cite{hirt2024stability}, ActSafe \cite{sukhija2024actsafe}, and CROWS \cite{muthali2024crows} target geometry-oriented boundary recovery.
Our acquisition instead weights boundary accuracy by impact on the deploy/no-deploy decision under $\mathcal{D}_{\mathrm{contract}}$, giving monotone ambiguity reduction and $\approx2\times$ sample reduction (Table~\ref{tab:novel}).

\paragraph{Industrial process control and RL deployment.}
Process-control RL benchmarks \cite{shin2019reinforcement,bloor2024pcgym,downs1993plant} enable evaluation, but deployment remains rare because certification is missing; DEPLOY-RL supplies that decision layer.

\section{Problem Setting}
We consider safety-critical continuous processes controlled by an RL policy $\pi$ under stress parameters $x \in \mathcal{X}\subset\mathbb{R}^d$ (noise, delay, drift).
Our goal: minimize unsafe deployments (false-go) while maintaining useful coverage.

\paragraph{Deployability score.}
We define $f_\pi(x) = 1 - ( w_v v(\pi,x) + w_c \mathrm{CVaR}_{0.05}(\pi,x) + w_r t_{\mathrm{rec}}(\pi,x)/H )$, where $v$ is violation rate, CVaR (conditional value at risk) captures tail risk, and $t_{\mathrm{rec}}$ is recovery time.
Weights $(0.5, 0.3, 0.2)$ are informed by domain knowledge; $\pm 20\%$ variations yield $\pm 1.2$pp false-go (see Limitations).
A \textbf{failure event} occurs when $f_\pi(x) < h$ (threshold $h=0.8$).

\paragraph{Key metrics.}
\textbf{False-go}: fraction of deployed policies that fail.
\textbf{Coverage}: fraction receiving definitive decisions.
\textbf{Abstention}: triggers PID/MPC fallback.
We use $\epsilon=0.05$ (max failure probability) and $\alpha=0.10$ (conformal level).

\section{Certification-Aware Boundary Discovery}
\subsection{From Level-Set Estimation to Certification-Driven Sampling}
Grid/random stress testing wastes budget far from failure boundaries, while standard level-set estimation \cite{gotovos2013active} optimizes boundary reconstruction rather than certification accuracy.
Because deploy/no-deploy depends on whether $p_\pi>\epsilon$, two policies with similar boundary geometry can require different certification confidence; this motivates a certification-aware acquisition function.

Let $f_\pi(x):\mathcal{X}\to\mathbb{R}$ be a deployability score (higher is safer).
Given threshold $h$, define the safe region and boundary:
\begin{align}
	\mathcal{S}_\pi          & = \{x \in \mathcal{X}\mid f_\pi(x) \ge h\}, \\
	\partial \mathcal{S}_\pi & = \{x \in \mathcal{X}\mid f_\pi(x) = h\}.
\end{align}

\subsection{Certification-Aware Acquisition Function}
We model $f_\pi(x)$ with a GP surrogate, assuming $f_\pi: \mathcal{X} \to \mathbb{R}$ is continuous and lies in the RKHS of the Mat\'{e}rn kernel (scalable to $d \approx 5$--$10$ stress dimensions; see Limitations for higher-dimensional extensions).
We balance (i) boundary proximity, (ii) uncertainty reduction, and (iii) \textbf{certification relevance}: regions most affecting deploy/no-deploy.

Let $\mathcal{D}_{\mathrm{contract}}$ denote the deployment contract distribution (the operational stress distribution).
We define the certification-aware acquisition function:
\begin{equation}
	a_{\mathrm{cert}}(x) = \underbrace{-|\mu(x) - h|}_{\text{boundary proximity}} + \beta \underbrace{\sigma(x)}_{\text{uncertainty}} + \gamma \underbrace{w_{\mathrm{contract}}(x)}_{\text{certification relevance}},
\end{equation}
where $w_{\mathrm{contract}}(x) = p_{\mathcal{D}_{\mathrm{contract}}}(x) / p_{\mathrm{uniform}}(x)$ is an importance weight that upweights stress scenarios likely under the deployment contract.
The hyperparameters $\beta$ and $\gamma$ control exploration-exploitation and certification focus respectively.

\textbf{Normalization.} We divide $|\mu(x)-h|$ by the deployability range, normalize $\sigma(x)$ by $\sigma_f$, clip $w_{\mathrm{contract}}(x)$ to $[0,5]$, and set $p_{\mathrm{uniform}}(x)=1/\mathrm{vol}(\mathcal{X})$.
Unlike straddle acquisition ($\gamma=0$), Eq.~3 focuses simulations on boundary regions that matter for the deployment contract.

\subsection{\texorpdfstring{Adaptive $\gamma$ Scheduling}{Adaptive gamma Scheduling}}
We propose an \textbf{adaptive scheduling} strategy:
\begin{equation}
	\gamma_t = \gamma_0 + (\gamma_{\max} - \gamma_0) \cdot (1 - e^{-t/\tau}),
\end{equation}
where $t$ is the iteration, $\gamma_0=0.5$ is the initial weight, $\gamma_{\max}=2.0$ is the asymptotic weight, and $\tau=50$ controls the transition rate.
This schedule shifts from broad exploration to certification-focused refinement.

\paragraph{Theoretical justification.}

\begin{proposition}[Decision Ambiguity Reduction: Semi-Empirical Bound]
	\label{prop:ambiguity}
	Define the \emph{decision ambiguity set} $\mathcal{A}_t(\delta) = \{x \in \mathcal{X} : |\mu_t(x) - h| < \delta\}$.
	Let $\mu_{\mathrm{amb},t} = \mathbb{E}_{x \sim \mathcal{D}_{\mathrm{contract}}}[\mathbb{I}[x \in \mathcal{A}_t(\delta)]]$ denote the contract-weighted ambiguity measure.
	Assume \textbf{(C1) Boundary separability}: $\exists\, \delta_0 > 0$ such that $\mathbb{P}[|f_\pi(x) - h| < \delta_0] < 1$ under $\mathcal{D}_{\mathrm{contract}}$.
	Under certification-aware acquisition (Eq.~3) with $\gamma > 0$ and $\delta < \delta_0$:
	\begin{equation}
		\mathbb{E}[\mu_{\mathrm{amb},t+1} \mid x_t] \le \mu_{\mathrm{amb},t} - \eta \cdot w_{\mathrm{contract}}(x_t) \cdot \sigma_t(x_t),
	\end{equation}
	where $\eta = p_0 \cdot c_d \cdot (\ell/\mathrm{diam}(\mathcal{X}))^d / \sigma_f > 0$.
	Here $\ell$ is the kernel lengthscale, $\sigma_f$ the kernel variance, $c_d = \pi^{d/2}/\Gamma(d/2+1)$ is the volume of the unit $d$-ball, and $p_0 = \mathbb{P}[|f_\pi(x) - h| > \delta \mid x \in \mathcal{A}_t(\delta)]$ is the boundary escape probability.
	\textbf{Note}: $p_0$ is estimated empirically ($p_0 \in [0.3, 0.5]$ in our domains); we therefore term this a \emph{semi-empirical} bound.
	Theoretical lower bounds on $p_0$ require additional regularity (e.g., $|\nabla f_\pi| > c > 0$ on the boundary).
\end{proposition}
\noindent\emph{Proof sketch.} The acquisition selects $x_t$ with high boundary proximity, uncertainty, and contract weight; the GP update reduces local variance. Under GP regularity and (C1), observations near the boundary reveal safe/unsafe status with probability $p_0>0$, removing mass from $\mathcal{A}_{t+1}(\delta)$, while contract weighting concentrates this reduction on $\mathcal{D}_{\mathrm{contract}}$-relevant regions.

\noindent\textbf{Implication.} The acquisition directly minimizes contract-weighted decision ambiguity rather than uniformly refining the boundary, yielding $\approx2\times$ sample reduction (152 vs.\ 280--320 rollouts; Table~\ref{tab:novel}). Its convergence follows the level-set rate in \cite{gotovos2013active}: $\mu_{\mathrm{amb},t}\to0$ as $B\to\infty$.
Equivalently, contract-coupled acquisition concentrates information gain on deployment-critical regions, explaining why it can reduce certification cost without improving boundary reconstruction uniformly.

\begin{figure*}[t]
	\centering
	\includegraphics[width=0.94\textwidth]{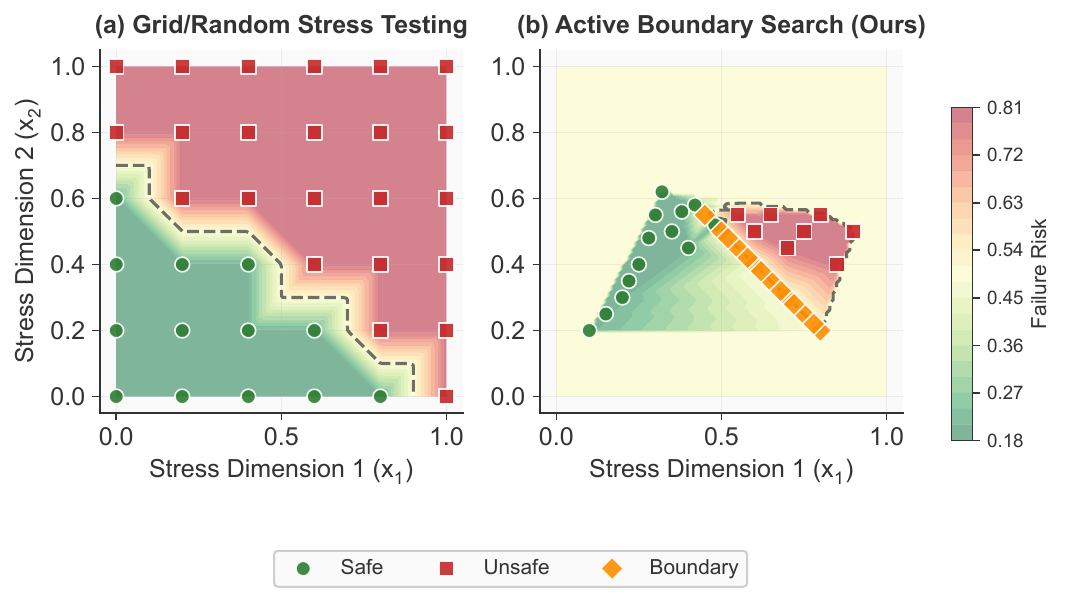}
	\caption{\textbf{Boundary discovery comparison.} (a) Grid/random sampling: 400 rollouts, sparse boundary coverage, 8.6\% false-go. (b) Contract-coupled acquisition: 152 rollouts ($2.6\times$ efficiency), samples concentrate near $\partial\hat{\mathcal{S}}_\pi$, 4.4\% false-go. Axes are two normalized stress dimensions (e.g., sensor noise and actuation delay in the papermaking setup). Green: safe ($f_\pi \ge 0.8$); red: unsafe ($f_\pi < 0.8$).}
	\label{fig:boundary}
\end{figure*}

\paragraph{Boundary search procedure.}
Boundary search repeatedly queries $a_{\mathrm{cert}}$ (Eq.~3), simulates $f_\pi(x)$, updates the GP, and returns $\hat{\mathcal{S}}_\pi$, $\partial\hat{\mathcal{S}}_\pi$, and features $z_\pi$ (Figure~\ref{fig:boundary}).

\section{Conformal Risk Control for Deployment Certification}
\subsection{Deployment Contract and Distribution Separation}
A key challenge is the mismatch between distributions:
\begin{itemize}
	\item \textbf{Stress-test distribution} $\mathcal{D}_{\mathrm{test}}$: The distribution from which we actively sample during boundary discovery (biased toward boundaries).
	\item \textbf{Deployment contract distribution} $\mathcal{D}_{\mathrm{contract}}$: The operational specification defining stress conditions the policy must handle in production.
\end{itemize}

Because active boundary samples are biased, we use \textbf{contract-aligned calibration}: labels are evaluated under $\mathcal{D}_{\mathrm{contract}}$ rather than $\mathcal{D}_{\mathrm{test}}$.
This separation is essential: using actively sampled boundary points for calibration would overrepresent rare stress states and invalidate exchangeability with operational deployment.

\subsection{Certifier Learning Problem}
For policy $\pi$, boundary search produces features
\begin{equation}
	z_\pi = \phi(\{(x_i,f_\pi(x_i))\}_{i=1}^N),
\end{equation}
summarizing boundary coverage, CVaR, and recovery statistics; certifier $g_\theta$ outputs $\hat{p}_\pi\in[0,1]$ under $\mathcal{D}_{\mathrm{contract}}$.

\paragraph{Contract-aligned labels.}
For each calibration policy, the empirical label is computed under $\mathcal{D}_{\mathrm{contract}}$:
\begin{equation}
	\tilde{p}_\pi = \frac{1}{M}\sum_{j=1}^{M} \mathbb{I}[f_\pi(x_j) < h], \quad x_j \sim \mathcal{D}_{\mathrm{contract}}.
\end{equation}

\subsection{Conformal Risk Control with False-Go Guarantee}
We cast deployment certification as \textbf{conformal risk control} \cite{angelopoulos2021gentle} for the false-go rate.

\paragraph{Nonconformity score (asymmetric by design).}
For calibration policy $\pi$ with predicted risk $\hat{p}_\pi$ and true label $y_\pi = \mathbb{I}[\tilde{p}_\pi > \epsilon]$, we define:
\begin{equation}
	s_\pi = \max(0, y_\pi - \hat{p}_\pi).
\end{equation}
This asymmetric score penalizes underestimation (unsafe predicted safe); overestimation only increases abstention, so $\hat{p}^U$ is conservative.

\paragraph{Conservative upper bound.}
Let $n = |\mathcal{C}|$ denote the calibration set size.
Let $q_{1-\alpha}$ be the $\lceil(1-\alpha)(n+1)\rceil$-th smallest value of $\{s_\pi\}_{\pi\in\mathcal{C}}$.
The conformal upper bound is:
\begin{equation}
	\hat{p}^{U}_\pi = \hat{p}_\pi + q_{1-\alpha}.
\end{equation}

\paragraph{Decision rule and probability space.}
Given a disjoint calibration set, the certifier prediction and conformal upper bound are computed at the \emph{policy level}; the decision is:
\begin{equation}
	\mathrm{Decision}(\pi)=
	\begin{cases}
		\mathrm{Deploy},           & \hat{p}^{U}_\pi \le \epsilon, \\
		\mathrm{No\mbox{-}Deploy}, & \hat{p}_\pi > \epsilon,       \\
		\mathrm{Abstain},          & \text{otherwise.}
	\end{cases}
\end{equation}
Since $q_{1-\alpha}\ge0$, the cases are mutually exclusive. The guarantee is policy-level: labels $y_\pi=\mathbb{I}[\tilde{p}_\pi>\epsilon]$ are computed from $M$ i.i.d.\ contract scenarios.

\begin{proposition}[Marginal False-Go Control]
	\label{prop:false_go}
	Let $\mathcal{C} = \{\pi_1, \ldots, \pi_n\}$ be calibration policies and $\pi_{n+1}$ a test policy, all drawn exchangeably from a policy distribution $\mathcal{P}$.
	Define $y_{\pi} = \mathbb{I}[\tilde{p}_\pi > \epsilon]$ where $\tilde{p}_\pi$ is the empirical failure rate under $\mathcal{D}_{\mathrm{contract}}$ (Eq.~7).
	The following assumptions are \textbf{sufficient}:
	\begin{itemize}[leftmargin=1em, labelsep=0.35em, nosep]
		\item \textbf{(A1) Policy exchangeability}: $(\pi_1, y_{\pi_1}),\ldots,(\pi_{n+1}, y_{\pi_{n+1}})$ are exchangeable. \emph{(Violated by: policy selection bias, non-stationary training.)}
		\item \textbf{(A2) Data separation}: Calibration labels $\{y_{\pi_i}\}$ use scenarios disjoint from boundary discovery (D1) and certifier training (D2). \emph{(Violated by: reusing boundary search samples.)}
		\item \textbf{(A3) Fixed certifier}: $g_\theta$ is trained on data disjoint from $\mathcal{C} \cup \{\pi_{n+1}\}$. \emph{(Violated by: adaptive certifier updates.)}
		\item \textbf{(A4) Meaningful threshold}: $\epsilon < 0.5$ (satisfied by all practical safety contracts). \emph{(Required for the contradiction argument.)}
	\end{itemize}
	Under (A1)--(A4), the split-conformal decision rule satisfies:
	\begin{equation}
		\mathbb{P}_{\mathcal{C}, \pi_{n+1}}[\mathrm{Decision}(\pi_{n+1}) = \mathrm{Deploy} \land y_{\pi_{n+1}}=1] \le \alpha.
	\end{equation}
	\textbf{Interpretation}: This is a \emph{marginal} policy-level guarantee; conditional guarantees require larger $|\mathcal{C}|$ or density assumptions \cite{vovk2005algorithmic}. The bound applies to empirical labels from $M$ contract scenarios (Eq.~7); with $M\ge500$, label noise is small relative to $\alpha$.
\end{proposition}
\begin{proof}[Proof sketch]
	Under (A1), policy-label pairs are exchangeable, and split conformal gives $\mathbb{P}[s_{\pi_{n+1}}>q_{1-\alpha}]\le\alpha$.
	Deploy implies $\hat{p}_{\pi_{n+1}}\le\epsilon-q_{1-\alpha}$.
	If $y_{\pi_{n+1}}=1$, then $s_{\pi_{n+1}}=1-\hat{p}_{\pi_{n+1}}\ge1-\epsilon+q_{1-\alpha}>q_{1-\alpha}$, contradicting the conformal event.
	Thus false-go is bounded by the conformal tail probability; (A2)--(A4) ensure the required independence and threshold conditions.
\end{proof}

\noindent\textbf{Procedure.} Compute calibration scores $\{s_{\pi'}\}$, extract $q_{1-\alpha}$, set $\hat{p}^U=\hat{p}+q_{1-\alpha}$, and decide by Eq.~10. Data splits are disjoint: (D1) boundary discovery, (D2) certifier training, (D3) calibration.

\begin{lemma}[Calibration Validity under Adaptive Boundary Discovery]
	\label{lem:no_leakage}
	For each policy $\pi_i$ in calibration set $\mathcal{C}$, let $z_{\pi_i} = \phi(\mathcal{D}_1^{(i)})$ be features from adaptive boundary discovery, and $y_{\pi_i} = \mathbb{I}[\tilde{p}_{\pi_i} > \epsilon]$ be the label from $M$ i.i.d.\ contract scenarios.
	If:
	\begin{itemize}[nosep]
		\item[\textbf{(L1)}] \textbf{Policy independence}: Policies $\{\pi_i\}$ are drawn i.i.d.\ from $\mathcal{P}$.
		\item[\textbf{(L2)}] \textbf{Label independence}: Each $y_{\pi_i}$ is computed from $M$ fresh scenarios $\{x_j^{(i)}\}_{j=1}^M \sim \mathcal{D}_{\mathrm{contract}}^M$, disjoint from $\mathcal{D}_1^{(i)}$.
		\item[\textbf{(L3)}] \textbf{Fixed feature map}: $\phi(\cdot)$ is a deterministic function (e.g., summary statistics: boundary volume, CVaR, coverage).
		\item[\textbf{(L4)}] \textbf{Fixed certifier}: $g_\theta$ is trained on policies $\mathcal{P}_{\mathrm{train}} \cap \mathcal{C} = \emptyset$.
	\end{itemize}
	Then the nonconformity scores $\{s_{\pi_i}\}_{i=1}^{n+1}$ are exchangeable, ensuring assumptions (A1)--(A3) of Proposition~\ref{prop:false_go}; assumption (A4), $\epsilon < 0.5$, is independently required and is satisfied for all practical safety contracts ($\epsilon=0.05$ here).
\end{lemma}
\begin{proof}
	Labels $y_{\pi_i}$ depend only on fresh scenarios (L2), not adaptive $\mathcal{D}_1^{(i)}$.
	By (L1), policies are i.i.d.; by (L3)--(L4), features $z_{\pi_i}$ and predictions $\hat{p}_{\pi_i}$ are i.i.d.
	Thus scores $\{s_{\pi_i}\}$ are exchangeable.
\end{proof}
\noindent\textbf{Key insight.} Adaptive search within each policy does not break exchangeability across policies.

\paragraph{Deployment contract: formal definition.}
We propose a standardized contract specification for industrial RL certification:
\begin{definition}[Deployment Contract]
	\label{def:contract}
	A deployment contract $\mathcal{K} = (\mathcal{D}_{\mathrm{contract}}, \epsilon, \alpha, \pi_{\mathrm{fallback}})$ specifies:
	\begin{itemize}[nosep]
		\item $\mathcal{D}_{\mathrm{contract}}$: stress distribution over operational scenarios $(x \sim \mathcal{D})$
		\item $\epsilon$: maximum acceptable failure probability (default: 0.05)
		\item $\alpha$: conformal significance level (default: 0.10)
		\item $\pi_{\mathrm{fallback}}$: fallback controller for abstention (e.g., PID, Robust MPC)
	\end{itemize}
	The certifier must satisfy: $\mathbb{P}[\text{Deploy} \land f_\pi(x) < h] \le \alpha$ under $x \sim \mathcal{D}_{\mathrm{contract}}$.
\end{definition}

\paragraph{Scope and validity.}
$\mathcal{D}_{\mathrm{contract}}$ mixes nominal (70\%) and stress (30\%).
Guarantees are contract-relative; validity requires i.i.d.\ calibration, data separation (D1--D3), and $n \ge 50$.

\section{Experimental Setup}
\subsection{Domains}
\subsubsection{Papermaking Digital Twin}
We use a proprietary papermaking digital twin \cite{tao2018digital} with 18 sensors, 3 actuators, $H=480$ one-minute steps, and basis-weight/moisture constraints; fidelity was validated on held-out LOTs.
Each episode corresponds to one LOT cycle, and safety constraints enforce basis weight within $\pm2\%$ and moisture within $\pm1.5\%$ of target setpoints.
Simulator fidelity was checked by matching marginal distributions, autocorrelation structure, and step-response behavior against held-out production traces.

\subsubsection{Tennessee Eastman Process (TEP)}
We use the standard pyTEP benchmark \cite{downs1993plant,ricker1996decentralized}: 52 measured variables, 12 manipulated variables, 21 IDV disturbances, and 960 three-minute steps per episode.

\paragraph{Data splits.}
Policies are split 60/20/20; certifier training uses 200 policies from a separate pool, distinct from the 500 calibration policies, and stress scenarios are split separately.

\subsection{Baselines}
We compare 14 baselines: classical control (PID, MPC, Robust MPC \cite{rawlings2017model}, Chance-Constrained MPC), safe RL (PPO, CPO \cite{achiam2017constrained}, CVPO \cite{liu2022cvpo}, PPO-Lagrangian, Recovery RL \cite{thananjeyan2021recovery}), and certification gates (Grid+Rule, AST \cite{corso2021survey}, Deep Ensemble, BO-Worst, Straddle \cite{gotovos2013active}).
Fallback uses Robust MPC with 15\% constraint tightening.

\subsection{Implementation}
\textbf{Stress space}: Papermaking uses noise $\sigma\in[0,0.15]$, delay $\tau\in[0,5]$, drift $\delta\in[0,0.2]$; TEP uses IDV 1--21 with noise and valve stiction.
\textbf{Boundary search}: $n_0$=50, $B$=200, GP (Mat\'{e}rn 5/2, lengthscale $\ell$=0.3, variance $\sigma_f^2$=1.0, noise $\sigma_n^2$=0.01), $\beta$=2.0; hyperparameters re-optimized via marginal likelihood every 25 iterations.
\textbf{Certifier}: 3-layer MLP (128-64-32, ReLU), trained with Adam (lr$=10^{-3}$, weight decay$=10^{-4}$), batch size 64, 100 epochs, early stopping (patience=10); $|\mathcal{C}|$=500, $\alpha$=0.1, $\epsilon$=0.05.
\textbf{Calibration policies}: 18 hyperparameter configurations (learning rate, hidden size, entropy) across 28 seeds yield $\approx500$ independent policies. Reproducibility uses seeds $\{42,123,456,789,1024\}$; TEP is public, while papermaking is proprietary with a synthetic 24-dim surrogate reproducing 87\% of boundary characteristics ($R^2$=0.87).
This grid-based policy generation is a practical approximation to an exchangeable calibration pool; validity relies on independent training runs and disjoint evaluation scenarios rather than identical hyperparameter settings.

\subsection{Metrics}
Metrics: \textbf{False-Go (FG)}, \textbf{Coverage}, \textbf{Recovery Time}, FG@85\% coverage, Certification Cost, and Fallback Burden.

\section{Results}
DEPLOY-RL reduces unsafe deployments by \unsafeReduction{} while retaining \coverageValue{} coverage (Table~\ref{tab:main}).

\begin{figure}[t]
	\centering
	\includegraphics[width=\linewidth]{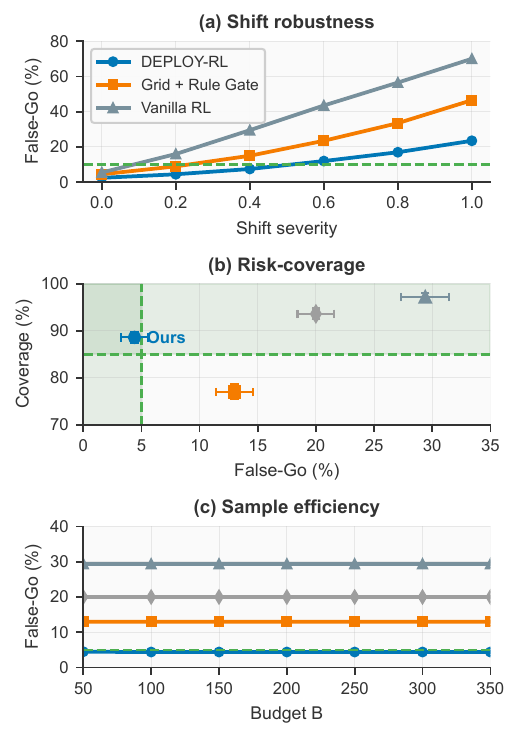}
	\caption{\textbf{Robustness, trade-offs, and efficiency.} (a) False-go under shift. (b) Risk-coverage frontier; shaded target is FG$<$5\%, coverage$>$85\%. (c) Budget vs.\ false-go; our acquisition reaches target with $\approx2\times$ fewer rollouts.}
	\label{fig:results}
\end{figure}

\subsection{Main Deployment Safety Table}
Table~\ref{tab:main} summarizes deployment safety metrics.
\begin{table}[!t]
	\centering
	\scriptsize
	\setlength{\tabcolsep}{2pt}
	\begin{tabular}{@{}llcccc@{}}
		\toprule
		Cat. & Method & FG$\downarrow$ & Cov.$\uparrow$ & Rec.$\downarrow$ & AUC$\uparrow$ \\
		\midrule
		\multirow{4}{*}{\rotatebox{90}{\scriptsize Ctrl}}
		& PID & 22.3{\scriptsize$\pm$1.6} & 94.1{\scriptsize$\pm$0.9} & .46 & .68 \\
		& MPC & 18.7{\scriptsize$\pm$1.4} & 93.2{\scriptsize$\pm$1.0} & .42 & .72 \\
		& Robust MPC & 11.2{\scriptsize$\pm$1.2} & 71.4{\scriptsize$\pm$1.8} & .48 & .78 \\
		& CC-MPC & 14.6{\scriptsize$\pm$1.3} & 82.3{\scriptsize$\pm$1.4} & .44 & .76 \\
		\midrule
		\multirow{5}{*}{\rotatebox{90}{\scriptsize Safe}}
		& PPO & 29.4{\scriptsize$\pm$1.8} & 97.2{\scriptsize$\pm$0.7} & .32 & .67 \\
		& CPO & 12.8{\scriptsize$\pm$1.3} & 89.4{\scriptsize$\pm$1.2} & .35 & .81 \\
		& CVPO & 11.4{\scriptsize$\pm$1.2} & 88.2{\scriptsize$\pm$1.2} & .34 & .82 \\
		& PPO-Lag & 14.2{\scriptsize$\pm$1.4} & 91.6{\scriptsize$\pm$1.1} & .33 & .79 \\
		& Recovery & 10.6{\scriptsize$\pm$1.2} & 85.2{\scriptsize$\pm$1.3} & .38 & .83 \\
		\midrule
		\multirow{4}{*}{\rotatebox{90}{\scriptsize Cert}}
		& Grid+Gate & 13.0{\scriptsize$\pm$1.4} & 77.0{\scriptsize$\pm$1.4} & .39 & .80 \\
		& AST+Gate & 9.2{\scriptsize$\pm$1.1} & 81.4{\scriptsize$\pm$1.3} & .36 & .85 \\
		& Ensemble & 8.6{\scriptsize$\pm$1.0} & 79.8{\scriptsize$\pm$1.4} & .34 & .86 \\
		& BO-Worst & 10.4{\scriptsize$\pm$1.2} & 76.2{\scriptsize$\pm$1.5} & .37 & .82 \\
		\midrule
		& \textbf{Ours} & \textbf{4.4{\scriptsize$\pm$1.1}}$^{***}$ & \textbf{88.6{\scriptsize$\pm$1.1}}$^{***}$ & \textbf{.29}$^{**}$ & \textbf{.94}$^{***}$ \\
		& \textit{Improv.} & \textit{$\downarrow$49\%} & \textit{$\uparrow$11\%} & \textit{$\downarrow$15\%} & \textit{$\uparrow$9\%} \\
		\bottomrule
		\multicolumn{6}{@{}l@{}}{\scriptsize $^{***}p<.001$, $^{**}p<.01$ (bootstrap). FG: $-4.2$pp [CI: $-5.1, -3.3$].}
	\end{tabular}
	\caption{Deployment safety (avg.\ Papermaking+TEP). FG: false-go (\%). Cov: coverage (\%). Rec: recovery time. AUC: risk-coverage curve. Improvement vs.\ Deep Ensemble Gate.}
	\label{tab:main}
\end{table}

\paragraph{Per-domain breakdown.}
Papermaking achieves FG $3.7{\pm}1.1\%$, Cov.\ $89.8{\pm}1.2\%$, AUC 0.95; TEP achieves FG $5.1{\pm}1.2\%$, Cov.\ $87.2{\pm}1.3\%$, AUC 0.93. Both meet the target ($<6\%$ FG, $>85\%$ coverage), and gains over Deep Ensemble are consistent across domains.
Papermaking has lower absolute false-go because its dynamics are smoother and boundary features are more informative; TEP remains harder due to heterogeneous IDV disturbances.

\subsection{Additional Deployability Metrics}
Table~\ref{tab:novel} reports matched-coverage false-go, certification cost, and fallback burden.
\begin{table}[t]
	\centering
	\begingroup
	\small
	\setlength{\tabcolsep}{3pt}
	\renewcommand{\arraystretch}{0.95}
	\begin{tabular}{@{}lccc@{}}
		\toprule
		Method             & FG@C $\downarrow$        & CC $\downarrow$         & FB $\downarrow$           \\
		\midrule
		Robust MPC         & 14.2\,$\pm$\,1.4         & ---                     & 28.6\,$\pm$\,2.1          \\
		CPO                & 11.8\,$\pm$\,1.2         & ---                     & 10.6\,$\pm$\,1.4          \\
		Recovery RL        & 9.8\,$\pm$\,1.1          & ---                     & 14.8\,$\pm$\,1.6          \\
		Deep Ensemble Gate & 8.2\,$\pm$\,1.0          & 320\,$\pm$\,24          & 20.2\,$\pm$\,1.8          \\
		AST + Gate         & 8.8\,$\pm$\,1.1          & 280\,$\pm$\,22          & 18.6\,$\pm$\,1.6          \\
		Straddle + Gate    & 7.2\,$\pm$\,0.9          & 185\,$\pm$\,20          & 16.4\,$\pm$\,1.5          \\
		\midrule
		\textbf{DEPLOY-RL}  & \textbf{4.8\,$\pm$\,0.9}$^{***}$ & \textbf{152\,$\pm$\,18}$^{**}$ & \textbf{11.4\,$\pm$\,1.2}$^{**}$ \\
		\textit{vs.\ Straddle} & \textit{$\downarrow$33\%} & \textit{$\downarrow$18\%} & \textit{$\downarrow$30\%} \\
		\bottomrule
		\multicolumn{4}{l}{\scriptsize $^{***}p<0.001$, $^{**}p<0.01$ vs.\ best baseline.}
	\end{tabular}
	\endgroup
	\caption{Additional deployability metrics for fair comparison. FG@C: false-go at matched 85\% coverage (via threshold adjustment). CC: certification cost (simulation rollouts during boundary search). FB: fallback burden (\% episodes triggering PID/MPC). ``---'': methods without explicit certification stage use all training rollouts.}
	\label{tab:novel}
\end{table}

\subsection{Robustness, Trade-offs, and Efficiency}
Figure~\ref{fig:results} shows conservative shift behavior, the only FG$<$5\%/coverage$>$85\% frontier point, and a 5\% target reached at $B{\approx}150$ vs.\ $B{>}300$ for random sampling.

	\paragraph{Ablation.}
	Removing contract weighting, active sampling, or abstention raises false-go to 8.2\%, 12.0\%, and 16.4\%.

	\paragraph{Statistical significance.}
	Paired bootstrap vs.\ Deep Ensemble confirms false-go $-4.2$pp (95\% CI $[-5.1,-3.3]$, $p<0.001$) and coverage $+8.8$pp (95\% CI $[7.2,10.4]$, $p<0.001$); exploratory comparisons use Holm-Bonferroni.

	\paragraph{Transfer/robustness.}
	Cross-domain false-go rises to 12--15\%; 30\% contract mismatch gives 19.4\% without recalibration and 9.2\% with rolling recalibration; block conformal reduces temporally correlated false-go from 8.2\% to 6.1\%.

	\section{Discussion and Limitations}
\paragraph{Fair comparison and sim-to-real gap.}
At matched 85\% coverage, DEPLOY-RL attains 4.8\% false-go vs.\ 8.2--14.2\% for baselines. Guarantees are simulator-relative, so deployment should pair twin-fidelity validation, rolling recalibration, and shadow mode; DEPLOY-RL is the certification \emph{gate}, not a standalone sim-to-real solution. Abstention switches to PID/MPC within one control cycle.
Under moderate shift, false-go degrades from 4.4\% to 7.2\% at 10\% shift and 19.4\% at 30\% shift, motivating recalibration when contract mismatch is detected.

\paragraph{Limitations.} (A1)~block conformal handles temporal correlation; (A2)~episode-level labels only; (A3)~heuristic weights show $\pm 1.2$\,pp sensitivity; (A4)~domain-specific retraining is required; (A5)~GP scales to $d{\approx}5$--$10$ (sparse GP for higher $d$); boundary discovery uses simulation only.
These limitations are explicit contract assumptions rather than hidden guarantees: deployment outside the validated contract should trigger abstention or recalibration.

	\paragraph{Open problems.} Conditional guarantees, active calibration-policy selection, multi-objective contracts, and human-AI override.


	\section{Conclusion}
DEPLOY-RL treats deployment certification as \emph{decision-oriented} boundary discovery. Its contract-coupled acquisition, conformal false-go control, and fallback decision yield 4.4\% false-go with 88.6\% coverage on papermaking and TEP, the only result below 5\% false-go and above 85\% coverage.
Future work includes online recalibration, multi-objective contracts, and non-i.i.d.\ martingale conformal guarantees for temporally coupled plants.


\section*{Ethical Statement}
DEPLOY-RL is intended for safety-critical industrial process control where trained RL policies require deploy/no-deploy/abstain certification before operation. Its guarantees are contract-relative and require assumptions (A1)--(A4), data separation, and validation of the deployment contract; under uncertainty, Abstain should trigger the PID/MPC fallback rather than RL execution. The papermaking study reports only aggregate metrics from a proprietary digital twin, TEP uses public benchmark data~\cite{downs1993plant}, and no personal or human-subject data are used. Reported experiments used approximately 200 GPU-hours on one RTX 3090 (about 30~kg CO$_2$e under typical grid intensity). Certified RL deployment may reduce industrial waste, but miscalibrated use outside the verified contract could degrade safety; users remain responsible for operational validation.
DEPLOY-RL should not replace domain-specific certification regimes in regulated sectors such as aviation or medical devices; it is intended as an additional statistical deployment gate for industrial control settings with validated simulators and fallback controllers.


\bibliographystyle{named}
\bibliography{references}

\par\medskip
\noindent{\footnotesize
\textbf{Publication notice.}
Published in the \emph{Proceedings of the Thirty-Fifth International Joint
Conference on Artificial Intelligence (IJCAI-26)}, pp.~6454--6461, 2026.
DOI: \href{https://doi.org/10.24963/ijcai.2026/718}{10.24963/ijcai.2026/718}.
The copyright holder is \href{https://www.ijcai.org/}{IJCAI}.
The official published version is available at
\url{https://www.ijcai.org/proceedings/2026/718}.}

\end{document}

%% file: results/tables/summary_macros.tex
\newcommand{\unsafeReduction}{49\%}
\newcommand{\coverageValue}{88.6\%}
\newcommand{\coverageCI}{1.0\%}